\documentclass{article}

\usepackage{microtype}
\usepackage{graphicx}
\usepackage{subfigure}
\usepackage{booktabs} %

\usepackage{hyperref}

\usepackage[accepted]{icml2025}

\usepackage{amsmath}
\usepackage{amssymb}
\usepackage{mathtools}
\usepackage{amsthm}

\usepackage[capitalize,noabbrev]{cleveref}

\theoremstyle{plain}

\theoremstyle{definition}

\theoremstyle{remark}

\usepackage[textsize=tiny]{todonotes}

\newcommand{\mypara}[1]{\textbf{\textit{#1}}.}

\icmltitlerunning{Align-then-Unlearn: Embedding Alignment for LLM Unlearning}

\begin{document}

\twocolumn[
\icmltitle{Align-then-Unlearn: Embedding Alignment for LLM Unlearning}

\icmlsetsymbol{equal}{*}

\begin{icmlauthorlist}
\icmlauthor{Philipp Spohn}{tum}
\icmlauthor{Leander Girrbach}{tum,mcmlmdsihelmholtz}
\icmlauthor{Jessica Bader}{tum,mcmlmdsihelmholtz}
\icmlauthor{Zeynep Akata}{tum,mcmlmdsihelmholtz}
\end{icmlauthorlist}

\icmlaffiliation{tum}{Technical University of Munich}
\icmlaffiliation{mcmlmdsihelmholtz}{Munich Center for Machine Learning, MDSI, Helmholtz Munich}

\icmlcorrespondingauthor{Philipp Spohn}{philipp.spohn@tum.de}

\icmlkeywords{Machine Learning, ICML}

\vskip 0.3in
]

\printAffiliationsAndNotice{}  %

\begin{abstract}
As large language models (LLMs) are trained on massive datasets, they have raised significant privacy and ethical concerns due to their potential to inadvertently retain sensitive information. Unlearning seeks to selectively remove specific data from trained models, such as personal information or copyrighted content.
Current approaches targeting specific output sequences at the token level often fail to achieve complete forgetting and remain susceptible to prompt rephrasing. We propose Align-then-Unlearn, a novel framework that performs unlearning in the semantic embedding space rather than directly on output tokens. Align-then-Unlearn first augments the LLM with an embedding prediction module trained to anticipate future context representations. Unlearning is then achieved by fine-tuning the model to minimize the similarity between these predicted embeddings and a target embedding that represents the concept to be removed. %
Initial results show that Align-then-Unlearn effectively removes targeted knowledge with minimal degradation in overall model utility.
These findings suggest that embedding-based unlearning offers a promising and robust approach to removing conceptual knowledge.
\end{abstract}
\section{Introduction}
\label{sec:introduction}
Large language models (LLMs) excel in a broad range of natural language processing tasks, thanks to their ability to learn rich semantic representations from vast text corpora. %
However, there is increasing concern regarding their retention of sensitive, private, or outdated information.
For instance, LLMs may inadvertently disclose personally identifiable information, such as email contacts or home addresses, or retain copyright-protected content.
These issues highlight the need for post-hoc removal of sensitive information from pre-trained models.
In this context, unlearning, the process of selectively removing specific data from a trained model, has emerged as a critical area of research.

\begin{figure}
    \centering
    \includegraphics[width=0.75\columnwidth]{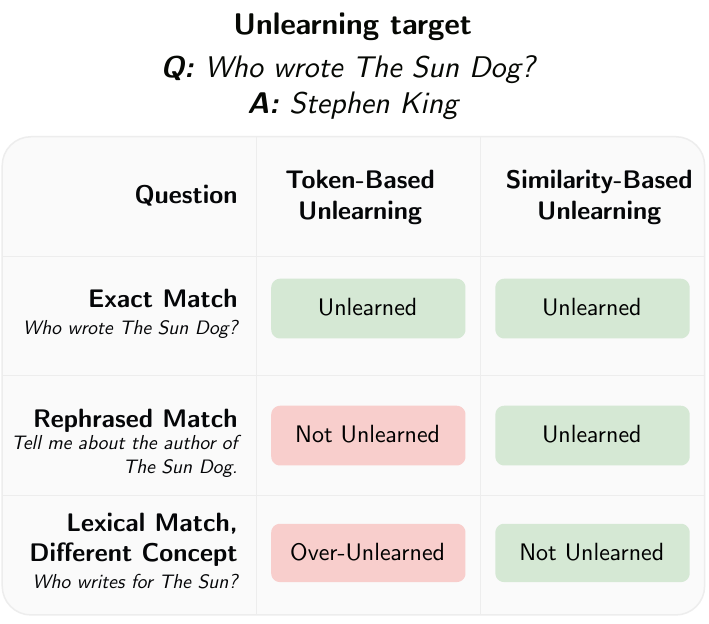}
    \vspace{-8px}
    \caption{Token-based unlearning fails on rephrasings and over-forgets unrelated knowledge. Align-then-Unlearn aims to address this by performing unlearning based on conceptual similarity.}
    \label{fig:teaser}
    \vspace{-15px}
\end{figure}

State-of-the-Art (SOTA) LLM unlearning methods typically remove knowledge via a specific set of text sequences, known as the forget set, targeting model outputs at the token level. However, this strategy presents challenges. First, operating in token space makes it difficult to precisely control \textit{what} is unlearned, as the unlearning target is defined only by the sequences in the forget set $D_f$, which can be quite large. Second, because the forget set includes a limited number of sequences, SOTA models often continue to expose the underlying semantic content in response to simple prompt rephrasings, even after unlearning, as illustrated in \autoref{fig:teaser}.

\begin{figure}
    \centering
    \includegraphics[width=0.8\columnwidth]{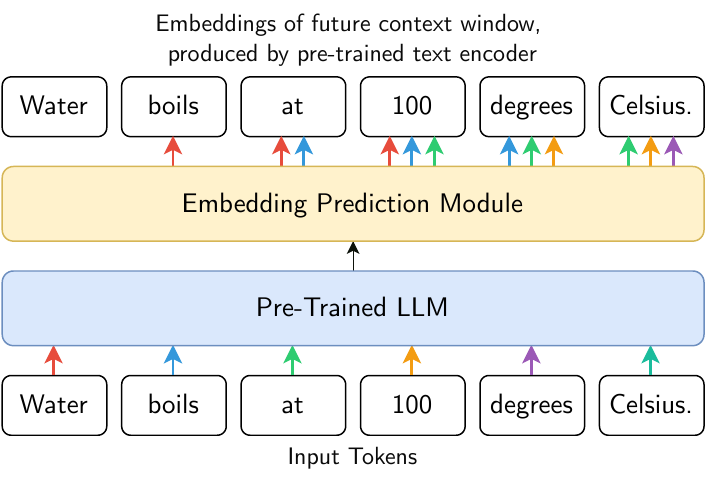}
    \caption{Align-then-Unlearn predicts a single embedding representing the next k tokens, rather than individual tokens. Each input token (e.g., ``Water'') maps to an embedding of its future k-token window (e.g., ``boils'', ``at'', ``100'', k=3), encouraging the model to capture conceptual meaning. %
    }
    \label{fig:embedding_pred}
\end{figure}

We propose Align-then-Unlearn, a framework addressing these limitations by operating in the embedding space. Unlike token-based methods, we capture conceptual meaning holistically by predicting embeddings representing multiple tokens at once, see \autoref{fig:embedding_pred}. This enables unlearning to operate over conceptual similarity, which could enable unlearning that is robust to rephrased prompts. Also, 
we target a single text embedding for unlearning, enabling more precise control over what is unlearned without the need for extensive dataset curation. We show that Align-then-Unlearn is competitive with SOTA methods in forgetting target information while maintaining overall model utility.

\section{Method}
\label{sec:method}
\begin{figure*}
    \centering
    \includegraphics[width=0.65\textwidth]{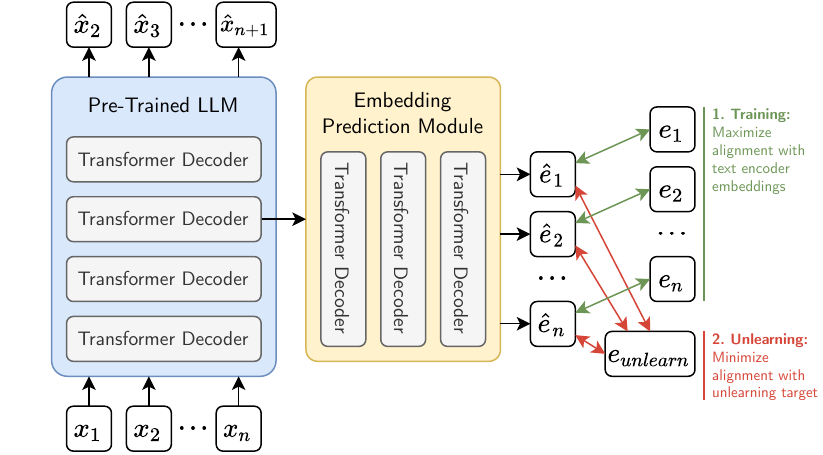}
    \caption{%
    In \textit{Align-then-Unlearn}, a pre-trained LLM is enhanced with an embedding prediction module that maps hidden states to future semantic embeddings. During training (green), predicted embeddings $\hat{e}_t$ are aligned with reference embeddings $e_t$ from a frozen text encoder. During unlearning (red), the model reduces alignment between $\hat{e}_t$ and a target concept embedding $e_{\text{unlearn}}$.}
    \label{fig:architecture}
\end{figure*}

Our method consists of two phases: \textit{alignment pre-training} and \textit{unlearning}. In the pre-training phase, an augmented LLM is trained to align its embedding predictions with those of a pre-trained text encoder. In the unlearning phase, the LLM is fine-tuned to minimize the similarity between the predicted embeddings and target unlearning concepts.

We augment the pre-trained LLM $M$, parameterized by $\theta_M$, with a small embedding prediction module $E$, parameterized by $\theta_E$, that learns to predict future semantic embeddings, as visualized in \autoref{fig:architecture}. Given an input token sequence $(x_1, \ldots, x_T)$, the LLM generates a sequence of hidden representations $(h_1, \ldots, h_T)$. At each position $t$, the prediction head $E$ maps the hidden state $h_t$ to a predicted embedding $\hat{e}_t = E(h_1, ..., h_t)$. Rather than predicting individual next tokens, each embedding $\hat{e}_t$ represents the meaning of the next $k$ tokens holistically, as represented in \autoref{fig:embedding_pred}.

\subsection{Alignment Pre-training}
\label{sec:embeddingprediction}

We train the embedding head using a loss that aligns predicted embeddings $\hat{e}_t$ with reference embeddings $e_t$, computed from the ground-truth future window $(x_{t+1}, \ldots, x_{t+k})$ with a frozen, pre-trained text encoder, such as MPNet \cite{song2020mpnet}.
The alignment loss $\mathcal{L}_{\text{align}}$ minimizes cosine distance between predicted and reference embeddings:
\begin{equation}
    \mathcal{L}_{\text{align}} = 1 - \operatorname{sim}(\hat{e}_t, e_t).
\end{equation}

where $\operatorname{sim}(a, b) = \frac{a \cdot b}{\|a\| \|b\|}$ denotes the cosine similarity. $\mathcal{L}_{\text{align}}$ measures the angular difference, commonly used in embedding spaces where angles reflect semantic similarity. We optimize the embedding predictor by:
\begin{equation}
    \theta_E^* = \arg\min_{\theta_E} \mathbb{E}_{x_{1:T}}\left[ \mathcal{L}_{\text{align}} \right].
\end{equation}
After training, the prediction head estimates the semantic meaning of likely next tokens.
\subsection{Unlearning}
\label{sec:unlearningmethod}
Once the embedding prediction module is trained, we unlearn by fine-tuning the LLM parameters to distance model outputs from sensitive concepts in embedding space. During this process, the embedding predictor $E$ remains frozen.

Given a concept description (e.g., ``Stephen King''), %
we compute its target embedding $e_{\text{unlearn}}$ using the frozen encoder, then fine-tune the LLM to minimize cosine similarity between predicted embeddings $\hat{e}_t$ and target embedding $e_{\text{unlearn}}$ via the unlearning loss:
\begin{equation}
    \mathcal{L}_{\text{unlearn}} = \max(0, \operatorname{sim}(\hat{e}_i, e_{\text{unlearn}}) - \tau),
\end{equation}
and we optimize:
\begin{equation}
    \theta_M^* = \arg\min_{\theta_M} \mathbb{E}_{x_{1:T}}\left[ \mathcal{L}_{\text{unlearn}} \right].
\end{equation}

Here, $\tau$ is a margin threshold. Only predictions that are sufficiently aligned with the target concept are penalized, thereby mitigating model degradation on unrelated tasks.

In practice, a single unlearning step is often insufficient, as the cosine similarity can quickly fall below threshold $\tau$. Once the similarity drops below $\tau$, the loss becomes zero, and unlearning is halted. To avoid early convergence, we alternate between realigning the prediction head and continuing unlearning in the main model.

\section{Analysis of Robustness}

The Align-then-Unlearn framework alternates between two objectives: the LLM ($M$) minimizes the unlearning loss $\mathcal{L}_{\text{unlearn}}(\theta_M, \theta_E)$, reducing similarity with the unlearning embedding $e_{\text{unlearn}}$, while the embedding prediction head ($E$) is realigned to predict future embeddings from $M$'s updated representations: $ \theta_E^* = \arg\min_{\theta_E} \mathcal{L}_{\text{align}}(\theta_E, \theta_M^{\text{updated}})$. In this step, $\theta_M^{\text{updated}}$ are the newly updated parameters of $M$.

This iterative process creates an adversarial dynamic: $M$ updates to obscure the unlearning target from $E$, while $E$ improves its ability to predict future semantics from $M$’s states. Robust unlearning requires $M$ to alter its representations $M(x_1, \ldots, x_t)$ so that even an optimally realigned $E$ (i.e., one whose parameters $\theta_E$ minimize $\mathcal{L}_{\text{align}}$) produces predictions $\hat{e}_t = E(M(x_1, \ldots, x_t))$ with minimal similarity to $e_{\text{unlearn}}$.

This approach goes deeper than masking the output, removing the unlearned concept from the hidden representation and enabling deeper and more robust forgetting.
However, fully suppressing $e_{\text{unlearn}}$ may also suppress related concepts, reducing overall utility.
In the limit of convergence, the model would suppress all signals that could recover $e_{\text{unlearn}}$, including those useful for other concepts. This reduces model utility and suggests that full convergence is impractical. Instead, partial unlearning via early stopping or thresholded objectives (as in our margin~$\tau$) is preferable. 

\section{Experiments}
\label{sec:experiments}
We evaluate the feasibility and effectiveness of the proposed Align-then-Unlearn using the Real-World Knowledge Unlearning (RWKU) benchmark \cite{jin2024rwku}. %

\subsection{Setup}
We use the \texttt{Phi-3-mini-4k-instruct}\footnote{https://huggingface.co/microsoft/Phi-3-mini-4k-instruct} model \cite{abdin2024phi} as the basis for unlearning, and \texttt{all-mpnet-base-v2}\footnote{https://huggingface.co/sentence-transformers/all-mpnet-base-v2}, based on MPNet \cite{song2020mpnet}, as the pre-trained text encoder for generating reference embeddings during head training and the target embedding $e_{\text{unlearn}}$. The embedding prediction head $E$ is 6 layers with hidden dimension 768. Unlearning updates are performed using sequences from the RWKU forget corpus.

\subsection{Metrics}
We adopt RWKU benchmark metrics \cite{jin2024rwku}:

    \textit{\textbf{Forget Score $\downarrow$}}: Measures the remaining knowledge of the unlearning target. Assessed through Fill-in-the-Blank (FB), Question Answering (QA), and Adversarial Attack (AA) probes. Lower scores indicate more effective forgetting.
    
    \textit{\textbf{Neighbor Score $\uparrow$}}: Measures the retained knowledge about related concepts. Assessed via QA and FB tasks. Higher scores are desirable, indicating preservation of related knowledge.
    
    \textit{\textbf{Utility Score $\uparrow$}}: Measures the general capability of the model after unlearning. Higher scores indicate better retention of overall performance.

For further details, please refer to RWKU benchmark paper \cite{jin2024rwku}.

\subsection{Unlearning Performance}

We compare Align-then-Unlearn with SOTA unlearning methods. As unlearning difficulty varies across targets, we adopt a dynamic training strategy where the unlearning threshold is gradually decreased. 
The RWKU benchmark reports one result per method, though extended training can further unlearn at the cost of utility. To make this trade-off transparent, we report results at the first checkpoint where the average accuracy on forget tasks drops below 50\%, 35\%, and 20\%. 
The results in \autoref{tab:unlearning_results} show that Align-then-Unlearn effectively removes target information, reducing Forget QA accuracy to 13.5\% while preserving overall model utility and still achieving 64.5\% accuracy on MMLU \cite{hendrycks2020measuring}.

\begin{table}[h]
\centering
\begin{small}
\begin{sc}
\resizebox{\columnwidth}{!}{
\begin{tabular}{lccc|cc|c}
\toprule
Method & \multicolumn{3}{c}{Forget $\downarrow$} & \multicolumn{2}{c}{Neighbor $\uparrow$} & Utility $\uparrow$ \\
 & FB & QA & AA & QA & FB & MMLU \\
\midrule
Before Unlearning                      & 47.1 & 47.4 & 55.8 & 61.4 & 56.2 & 64.4 \\
ICU{\scriptsize~\cite{pawelczyk2023context}}        & 45.2 & 34.6 & 32.2 & 56.1 & 52.9 & 63.9 \\
GA* (F){\scriptsize~\cite{jang2022knowledge}}       & 37.1 & 37.9 & 46.4 & 59.2 & 51.8 & 64.4 \\
GA* (L){\scriptsize~\cite{jang2022knowledge}}       & 46.2 & 47.5 & 55.8 & 61.2 & 55.1 & 64.4 \\
GA (F){\scriptsize~\cite{jang2022knowledge}}        & 17.8 & 14.3 & 26.3 & 51.7 & 49.7 & 64.3 \\
GA (L){\scriptsize~\cite{jang2022knowledge}}        & 40.5 & 37.8 & 49.5 & 60.1 & 55.2 & 64.2 \\
DPO (F){\scriptsize~\cite{rafailov2023direct}}      & 25.0 & 19.1 & 29.9 & 39.6 & 41.4 & 63.0 \\
DPO (L){\scriptsize~\cite{rafailov2023direct}}      & 44.1 & 45.6 & 54.9 & 60.5 & 56.2 & 64.3 \\
NPO (F){\scriptsize~\cite{zhang2024negative}}       & 22.5 & 16.9 & 27.3 & 53.6 & 50.5 & 64.2 \\
RT (F){\tiny~\cite{ishibashi2023knowledge}}   & 47.6 & 46.6 & 55.4 & 61.5 & 57.2 & 64.1 \\
\midrule
ATU (Ours; $50\%$) & 36.3 & 40.5 & 48.8 & 64.4 & 55.8 & 64.2 \\
ATU (Ours; $35\%$) & 24.1 & 24.8 & 37.2 & 56.4 & 51.3 & 64.8 \\
ATU (Ours; $20\%$) & 13.5 & 15.3 & 25.9 & 52.3 & 41.5 & 64.5 \\
\bottomrule
\end{tabular}
}
\end{sc}
\end{small}
\caption{Unlearning results %
(average of 15 targets). 
Baselines follow from the RWKU benchmark \cite{jin2024rwku}, are either Full (F) or LoRA (L) fine-tuned, and * denotes training on pseudo-ground truth forget corpus.
Evaluation with Fill-in-the-Blank (FB), Question Answering (QA), and Adversarial Attack (AA) probes.}
\label{tab:unlearning_results}

\end{table}

\subsection{Forgetting / Model Utility Trade-Off}
\begin{figure}
    \centering
    \includegraphics[width=\linewidth]{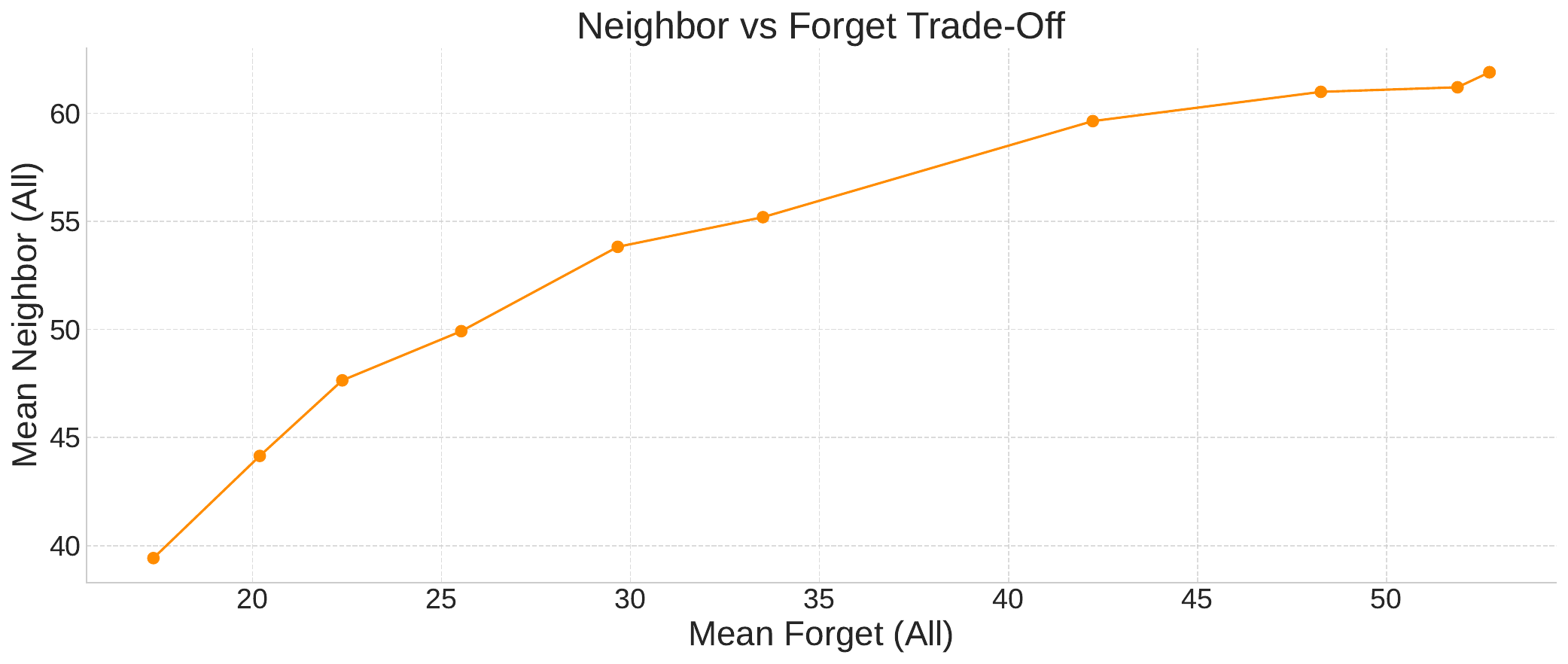}
    \caption{Trade-off between effective forgetting and retention of related knowledge. The results are averaged across 15 unlearning targets; metrics computed at corresponding unlearning steps.}
    \label{fig:trade_off}
\end{figure}

To analyze the forgetting–performance trade-off, we plot the average forget accuracy against the accuracy on neighboring knowledge across unlearning stages. As \autoref{fig:trade_off} demonstrates, stronger forgetting (lower forget scores)  typically reduces nearby concept knowledge.

\subsection{Optimal Layer for Unlearning}
Our method enables unlearning at any hidden layer of the model, allowing finer control over where conceptual suppression is applied.

\begin{figure}[h]
    \centering
    \includegraphics[width=0.48\linewidth]{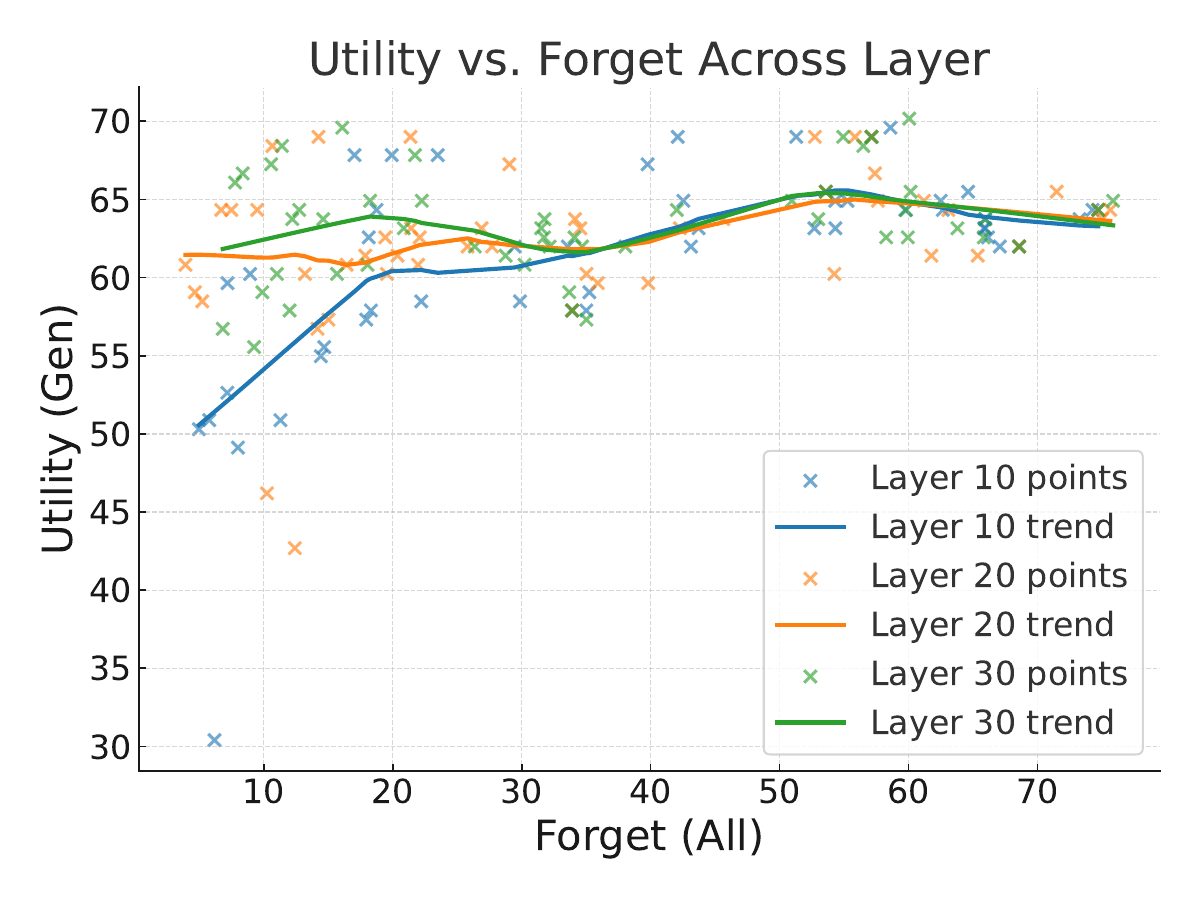}
    \includegraphics[width=0.48\linewidth]{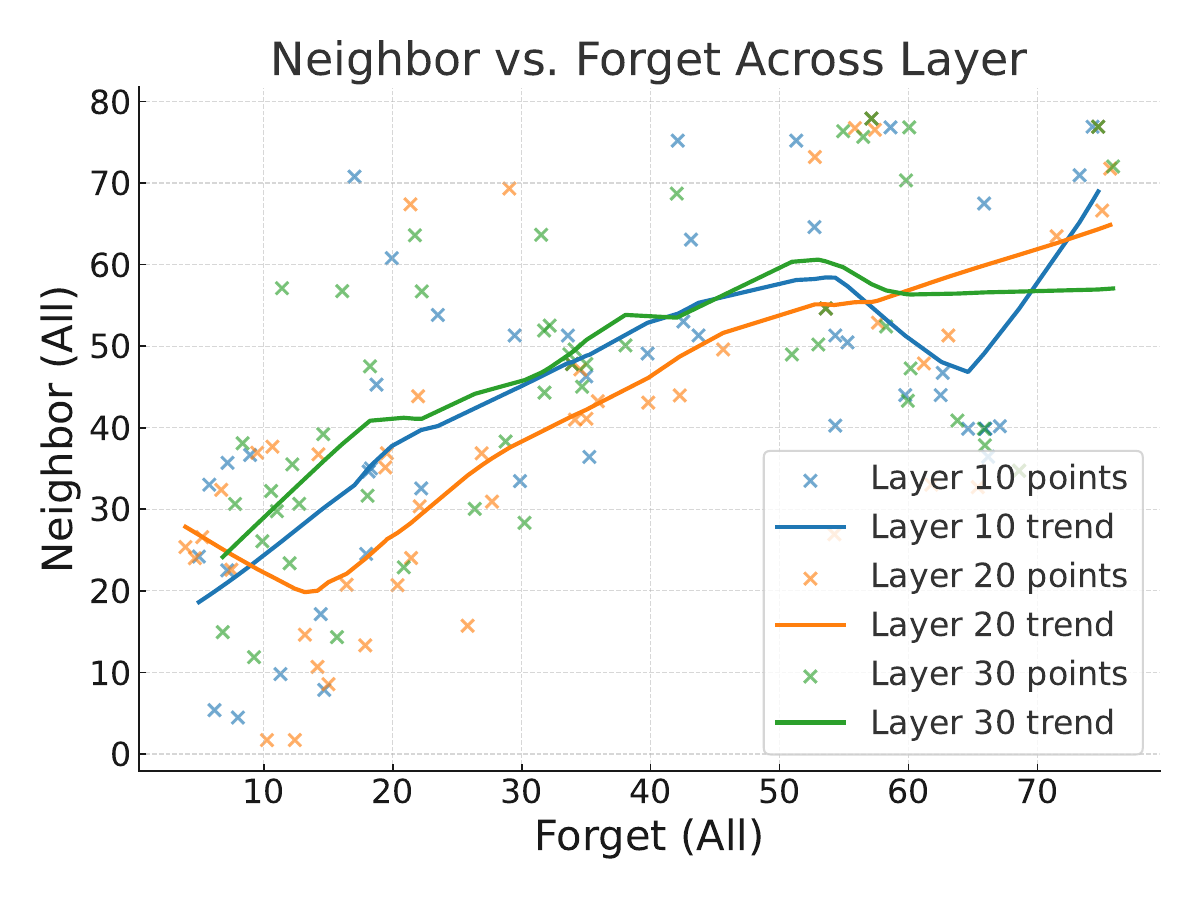}
    \caption{Trade-off between forgetting and model performance. Left: Utility vs. Forget. Right: Neighbor Score vs. Forget. Results averaged across 5 targets for each layer.}
    \label{fig:trade_off_optimal_layer}
\end{figure}

\begin{figure}[h]
    \centering
    \includegraphics[width=\linewidth]{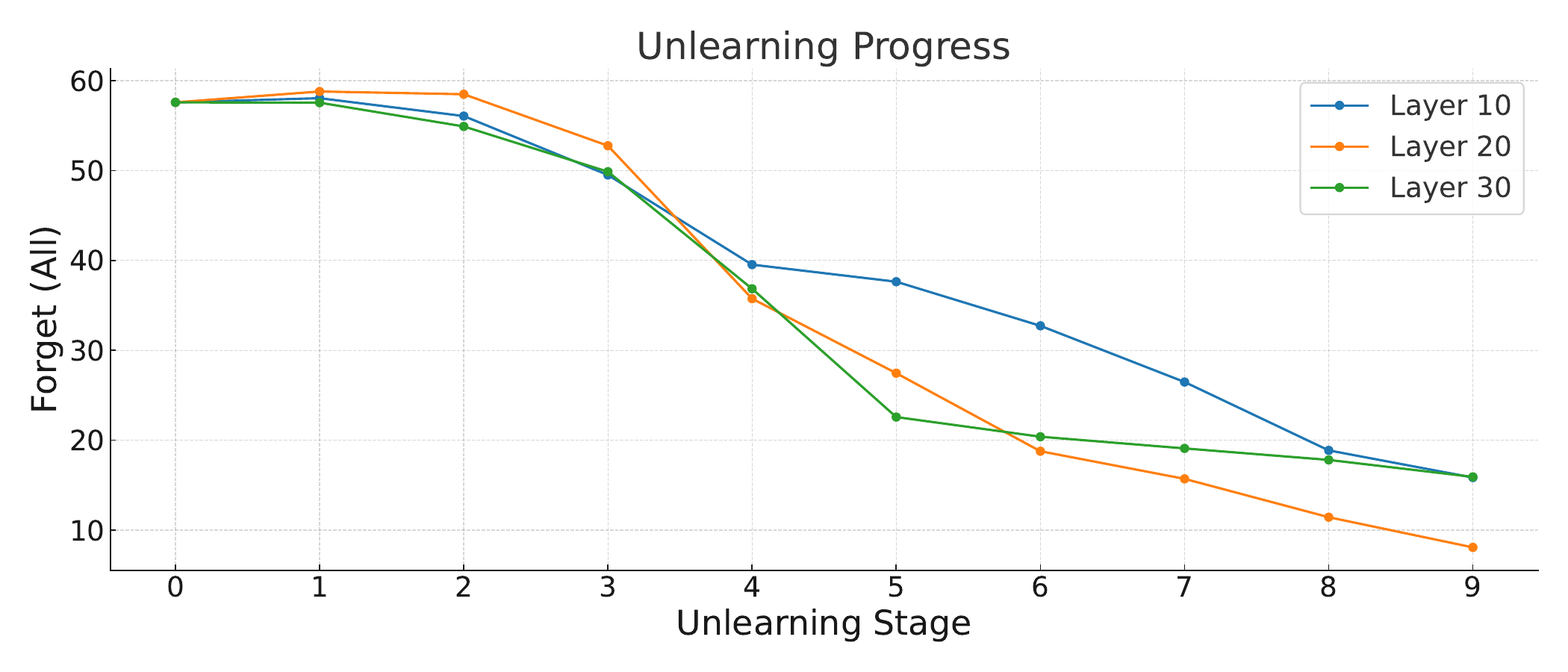}
    \caption{Comparison of forget scores over time when applying unlearning to layers 10, 20, and 30. Average of 5 targets.}
    \label{fig:unlearning_progress}
\end{figure}

We evaluated the effect of applying Align-then-Unlearn at layers 10, 20, and 30 of \texttt{Phi-3-mini-4k-instruct} (32 layers total). As shown in \autoref{fig:trade_off_optimal_layer} and \autoref{fig:unlearning_progress}, no single layer consistently outperforms others; all show similar forgetting–utility trade-offs. However, individual targets exhibit clear differences. For instance, when unlearning ``Warren Buffett'', applying the objective at layer 10 resulted in a forget accuracy of 54.32\%, while applying it at layer 20 lowered it to just 12.40\%. This suggests unlearning effectiveness depends on where in the network the unlearning is applied, hinting at future research directions like target-specific layer selection or multi-layer strategies.

\section{Discussion \& Conclusion}
Our Align-then-Unlearn framework highlights the potential of embedding-based unlearning for LLMs, offering a novel approach that focuses on conceptual similarity rather than specific tokens to remove knowledge. By leveraging a pre-trained text encoder, this method enables data-efficient unlearning, while the threshold $\tau$ provides fine-grained control over the scope of unlearning. 

However, limitations remain. %
Tuning $\tau$ for a single target and applying it to others reduced effectiveness, suggesting the need for dynamic threshold adjustment.
Moreover, lost neighbor knowledge suggests that embedding-based unlearning may overgeneralize, inadvertently affecting related %
concepts.

\mypara{Future Work}
Future work could explore dynamic $\tau$ adjustment based on target characteristics and unlearning loss. The embedding prediction module may be applied to multiple or specific layers, and joint training of the encoder and embedding head should be investigated. Additionally, evaluating effectiveness for unlearning exact text sequences (e.g., passwords) could assess its suitability for such scenarios.

\section*{Impact Statement}
This work proposes a concept-based unlearning method for language models, enhancing robustness to rephrasing and aiding privacy compliance. While it enables precise forgetting, it also risks degrading related knowledge or enabling misuse for information suppression, highlighting the need to understand trade-offs for responsible deployment.

\section*{Code Availability}
Our code is publicly available at \url{https://github.com/ExplainableML/align-then-unlearn}.

\section*{Acknowledgments}
This work was partially funded by the ERC (853489 - DEXIM) and the Alfried Krupp von Bohlen und Halbach Foundation, which we thank for their generous support. The authors gratefully acknowledge the scientific support and resources of the AI service infrastructure \textit{LRZ AI Systems} provided by the Leibniz Supercomputing Centre (LRZ) of the Bavarian Academy of Sciences and Humanities (BAdW), funded by Bayerisches Staatsministerium für Wissenschaft und Kunst (StMWK).

\bibliography{main}

\begin{thebibliography}{27}
\providecommand{\natexlab}[1]{#1}
\providecommand{\url}[1]{\texttt{#1}}
\expandafter\ifx\csname urlstyle\endcsname\relax
  \providecommand{\doi}[1]{doi: #1}\else
  \providecommand{\doi}{doi: \begingroup \urlstyle{rm}\Url}\fi

\bibitem[Abdin et~al.(2024)Abdin, Aneja, Awadalla, Awadallah, Awan, Bach,
  Bahree, Bakhtiari, Bao, Behl, et~al.]{abdin2024phi}
Abdin, M., Aneja, J., Awadalla, H., Awadallah, A., Awan, A.~A., Bach, N.,
  Bahree, A., Bakhtiari, A., Bao, J., Behl, H., et~al.
\newblock {Phi-3} technical report: A highly capable language model locally on
  your phone.
\newblock In \emph{arXiv}, 2024.

\bibitem[Carlini et~al.(2023)Carlini, Ippolito, Jagielski, Lee, Tramer, and
  Zhang]{carlini2022quantifying}
Carlini, N., Ippolito, D., Jagielski, M., Lee, K., Tramer, F., and Zhang, C.
\newblock Quantifying memorization across neural language models.
\newblock In \emph{ICLR}, 2023.

\bibitem[Caves~Jr \& Carus(2014)Caves~Jr and Carus]{caves2014future}
Caves~Jr, J.~P. and Carus, W.~S.
\newblock The future of weapons of mass destruction: their nature and role in
  2030, 2014.

\bibitem[Eldan \& Russinovich(2023)Eldan and Russinovich]{eldan2023s}
Eldan, R. and Russinovich, M.
\newblock Who’s {Harry Potter}? approximate unlearning for {LLMs}, 2023.

\bibitem[Freeman et~al.(2024)Freeman, Rippe, Debenedetti, and
  Andriushchenko]{freeman2024exploring}
Freeman, J., Rippe, C., Debenedetti, E., and Andriushchenko, M.
\newblock Exploring memorization and copyright violation in frontier {LLMs}: A
  study of the {New York Times} v. {OpenAI} 2023 lawsuit.
\newblock In \emph{NeurIPS Workshop Safe Generative AI}, 2024.

\bibitem[Hendrycks et~al.(2021)Hendrycks, Burns, Basart, Zou, Mazeika, Song,
  and Steinhardt]{hendrycks2020measuring}
Hendrycks, D., Burns, C., Basart, S., Zou, A., Mazeika, M., Song, D., and
  Steinhardt, J.
\newblock Measuring massive multitask language understanding.
\newblock In \emph{ICLR}, 2021.

\bibitem[Hu et~al.(2025)Hu, Fu, Wu, and Smith]{hu2025unlearning}
Hu, S., Fu, Y., Wu, S., and Smith, V.
\newblock Unlearning or obfuscating? jogging the memory of unlearned llms via
  benign relearning.
\newblock In \emph{ICLR}, 2025.

\bibitem[Huang et~al.(2022)Huang, Shao, and Chang]{huang2022large}
Huang, J., Shao, H., and Chang, K. C.-C.
\newblock Are large pre-trained language models leaking your personal
  information?
\newblock In \emph{EMNLP (Findings)}, 2022.

\bibitem[Huang et~al.(2024)Huang, Zhou, Wang, Morstatter, Zhang, Poon, and
  Chen]{huang2024offset}
Huang, J.~Y., Zhou, W., Wang, F., Morstatter, F., Zhang, S., Poon, H., and
  Chen, M.
\newblock Offset unlearning for large language models.
\newblock In \emph{arXiv}, 2024.

\bibitem[Ishibashi \& Shimodaira(2023)Ishibashi and
  Shimodaira]{ishibashi2023knowledge}
Ishibashi, Y. and Shimodaira, H.
\newblock Knowledge sanitization of large language models.
\newblock In \emph{arXiv}, 2023.

\bibitem[Jang et~al.(2023)Jang, Yoon, Yang, Cha, Lee, Logeswaran, and
  Seo]{jang2022knowledge}
Jang, J., Yoon, D., Yang, S., Cha, S., Lee, M., Logeswaran, L., and Seo, M.
\newblock Knowledge unlearning for mitigating privacy risks in language models.
\newblock In \emph{ACL}, 2023.

\bibitem[Ji et~al.(2024)Ji, Liu, Zhang, Liu, Kompella, Liu, and
  Chang]{ji2024reversing}
Ji, J., Liu, Y., Zhang, Y., Liu, G., Kompella, R., Liu, S., and Chang, S.
\newblock Reversing the forget-retain objectives: An efficient llm unlearning
  framework from logit difference.
\newblock In \emph{NeurIPS}, 2024.

\bibitem[Jin et~al.(2024)Jin, Cao, Wang, He, Yuan, Li, Chen, Liu, and
  Zhao]{jin2024rwku}
Jin, Z., Cao, P., Wang, C., He, Z., Yuan, H., Li, J., Chen, Y., Liu, K., and
  Zhao, J.
\newblock {RWKU}: Benchmarking real-world knowledge unlearning for large
  language models.
\newblock In \emph{NeurIPS}, 2024.

\bibitem[Kim et~al.(2023)Kim, Yun, Lee, Gubri, Yoon, and Oh]{kim2023propile}
Kim, S., Yun, S., Lee, H., Gubri, M., Yoon, S., and Oh, S.~J.
\newblock {ProPILE}: Probing privacy leakage in large language models.
\newblock In \emph{NeurIPS}, 2023.

\bibitem[Maini et~al.(2024)Maini, Feng, Schwarzschild, Lipton, and
  Kolter]{maini2024tofu}
Maini, P., Feng, Z., Schwarzschild, A., Lipton, Z.~C., and Kolter, J.~Z.
\newblock {TOFU}: A task of fictitious unlearning for {LLMs}.
\newblock In \emph{COLM}, 2024.

\bibitem[Patil et~al.(2024)Patil, Hase, and Bansal]{patil2023can}
Patil, V., Hase, P., and Bansal, M.
\newblock Can sensitive information be deleted from {LLMs}? objectives for
  defending against extraction attacks.
\newblock In \emph{ICLR}, 2024.

\bibitem[Pawelczyk et~al.(2024)Pawelczyk, Neel, and
  Lakkaraju]{pawelczyk2023context}
Pawelczyk, M., Neel, S., and Lakkaraju, H.
\newblock In-context unlearning: Language models as few shot unlearners.
\newblock In \emph{ICML}, 2024.

\bibitem[Rafailov et~al.(2023)Rafailov, Sharma, Mitchell, Manning, Ermon, and
  Finn]{rafailov2023direct}
Rafailov, R., Sharma, A., Mitchell, E., Manning, C.~D., Ermon, S., and Finn, C.
\newblock {Direct Preference Optimization}: Your language model is secretly a
  reward model.
\newblock In \emph{NeurIPS}, 2023.

\bibitem[Scholten et~al.(2025)Scholten, Günnemann, and
  Schwinn]{scholten2025probabilistic}
Scholten, Y., Günnemann, S., and Schwinn, L.
\newblock A probabilistic perspective on unlearning and alignment for large
  language models.
\newblock In \emph{ICLR}, 2025.

\bibitem[Schwarzschild et~al.(2024)Schwarzschild, Feng, Maini, Lipton, and
  Kolter]{schwarzschild2024rethinking}
Schwarzschild, A., Feng, Z., Maini, P., Lipton, Z., and Kolter, J.~Z.
\newblock Rethinking {LLM} memorization through the lens of adversarial
  compression.
\newblock In \emph{NeurIPS}, 2024.

\bibitem[Seyito{\u{g}}lu et~al.(2024)Seyito{\u{g}}lu, Kuvshinov, Schwinn, and
  G{\"u}nnemann]{seyitouglu2024extracting}
Seyito{\u{g}}lu, A., Kuvshinov, A., Schwinn, L., and G{\"u}nnemann, S.
\newblock Extracting unlearned information from {LLMs} with {Activation
  Steering}.
\newblock In \emph{NeurIPS Workshop Safe Generative AI}, 2024.

\bibitem[Song et~al.(2020)Song, Tan, Qin, Lu, and Liu]{song2020mpnet}
Song, K., Tan, X., Qin, T., Lu, J., and Liu, T.-Y.
\newblock {MPNet}: Masked and permuted pre-training for language understanding.
\newblock In \emph{NeurIPS}, 2020.

\bibitem[Thaker et~al.(2024)Thaker, Maurya, Hu, Wu, and
  Smith]{thaker2024guardrail}
Thaker, P., Maurya, Y., Hu, S., Wu, Z.~S., and Smith, V.
\newblock Guardrail baselines for unlearning in {LLMs}.
\newblock In \emph{arXiv}, 2024.

\bibitem[Yao et~al.(2024)Yao, Xu, and Liu]{yao2024large}
Yao, Y., Xu, X., and Liu, Y.
\newblock Large language model unlearning.
\newblock In \emph{NeurIPS}, 2024.

\bibitem[Zhang et~al.(2024{\natexlab{a}})Zhang, Lin, Bai, and
  Mei]{zhang2024negative}
Zhang, R., Lin, L., Bai, Y., and Mei, S.
\newblock {Negative Preference Optimization}: From catastrophic collapse to
  effective unlearning.
\newblock In \emph{COLM}, 2024{\natexlab{a}}.

\bibitem[Zhang et~al.(2024{\natexlab{b}})Zhang, Wang, Li, Wu, Tang, Liu, He,
  Yin, and Wang]{zhang2024catastrophic}
Zhang, Z., Wang, F., Li, X., Wu, Z., Tang, X., Liu, H., He, Q., Yin, W., and
  Wang, S.
\newblock Catastrophic failure of {LLM} unlearning via quantization.
\newblock In \emph{arXiv}, 2024{\natexlab{b}}.

\bibitem[Zhong et~al.(2023)Zhong, Wu, Manning, Potts, and
  Chen]{zhong2023mquake}
Zhong, Z., Wu, Z., Manning, C.~D., Potts, C., and Chen, D.
\newblock {MQuAKE}: Assessing knowledge editing in language models via
  multi-hop questions.
\newblock In \emph{EMNLP}, 2023.

\end{thebibliography}
\bibliographystyle{icml2025}

\newpage
\appendix
\onecolumn
\section{Related Work}
\subsection{LLMs Memorize Sensitive Information}
LLMs may reveal personally identifiable information, such as email contacts or home addresses \cite{huang2022large,kim2023propile}, or memorize copyright-protected data \cite{carlini2022quantifying,schwarzschild2024rethinking,freeman2024exploring}.
These issues underscore the necessity to develop methods for post-hoc removal of sensitive information from a pre-trained model $\pi_{\theta}$.
\subsection{State-of-the-Art Unlearning Approaches}
\mypara{In-Context Unlearning}
In-context unlearning \cite{thaker2024guardrail} adjusts model behavior via prompts, typically without changing weights. However, it is susceptible to jailbreak attacks and cannot be used when model weights are accessible.

\mypara{Gradient-Ascent-Based Unlearning Methods}
One of the most straightforward strategies for unlearning involves directly applying gradient ascent to reduce a model’s performance on the data that should be forgotten \cite{jang2022knowledge}. This is done by maximizing the log-likelihood of the forget set $D_f = \{(x,y)\}$:
\begin{equation}
    \mathcal{L}_{\text{forget}}~=~\log~\pi_{\theta}(y \mid x).
\end{equation}

However, this method often leads to a significant deterioration in the model’s overall capabilities. To address this, gradient difference has been proposed \cite{maini2024tofu}, which aims to preserve the model’s performance on data that should be retained. This is typically achieved by minimizing the negative log-likelihood over the retain set:
\begin{equation}
    \mathcal{L}_{\text{retain}}~=\nobreak-\log~\pi_{\theta}(y \mid x).
\end{equation}

More advanced variants \cite{yao2024large,maini2024tofu} incorporate a KL divergence term to penalize deviation from the pre-unlearning model:
\begin{equation}
    \mathcal{L}_{\text{KL}}~=~\mathbb{E}_{(x, y) \sim D_r} \left[ \text{KL}(\pi_{\text{ref}}(y \mid x) \Vert \pi_{\theta}(y \mid x)) \right],
\end{equation}
where $\pi_{\text{ref}}$ denotes the model before any unlearning is applied.

Although incorporating retention loss can mitigate utility loss, performance degradation still tends to occur, especially as the size of the forget set increases \cite{maini2024tofu}.

\mypara{Preference-Optimization-Based Unlearning Methods}
Another set of approaches borrows ideas from Direct Preference Optimization (DPO) \cite{rafailov2023direct}, which uses pairs of responses ($y_{+}, y_{-}$) to a prompt $x$ to align a model with human preferences. For the purpose of unlearning, Negative Preference Optimization (NPO)~\cite{zhang2024negative} adapts this to work only with the negative example. The goal is to decrease the model's likelihood of generating $y_{-}$ given $x$, relative to a reference model (the model before unlearning):
\begin{equation}
    \mathcal{L}_{\text{NPO}, \beta} = \frac{2}{\beta} \mathbb{E}_{(x, y_{-}) \sim D_f} \left[ \log \left( 1 + \left( \frac{\pi_{\theta}(y_{-} \mid x)}{\pi_{\text{ref}}(y_{-} \mid x)} \right)^\beta \right) \right].
\end{equation}

Here, $\pi_{\theta}$ is the model being updated, $\pi_{\text{ref}}$ is the original reference model, and $\beta$ controls the balance between unlearning speed and model utility. NPO is more stable than gradient ascent methods because it reduces updates for already unlearned data. However, it still struggles to maintain utility with larger forget sets.

\mypara{Methods Based on Logit Differences}
Logit-difference approaches \cite{eldan2023s, ji2024reversing, huang2024offset} first train a \textit{reinforced} model to enhance its prediction performance on the forget set, effectively encouraging it to produce unwanted outputs. They subsequently use the logit differences between this enhanced model and the original model to guide the unlearning process.

\citet{eldan2023s} illustrate this by attempting to erase knowledge about specific concepts (e.g., ``Harry Potter'') via generic fine-tuning completions. They generate generic completions by adjusting logits:
\begin{equation}
    z_{\text{generic}} = z_{\text{original}} - \alpha \text{ReLU}(z_{\text{reinforced}} - z_{\text{original}}),
\end{equation}
where $z$ denotes logits and $\alpha$ is a hyperparameter.

A common element in all of the optimization-based methods is that they directly operate on the output tokens of the model, which may limit robustness to rephrased questions or harm the overall language abilities of the model.

\subsection{Evaluation Methods}
Evaluating unlearning involves measuring both how effectively a model forgets target information (forget quality) and how well it retains overall performance (model utility). The Weapons of Mass Destruction Proxy (WMDP) benchmark \cite{caves2014future} can be used to evaluate unlearning in sensitive domains. However, it is based on multiple-choice questions, which means it may miss leaked knowledge in open-ended questions. More rigorous methods include TOFU \cite{maini2024tofu}, which uses synthetic author profiles the model has been fine-tuned on. This means that the optimal unlearned model is available (the model before fine-tuning, which has never been trained on the forget set). This enables more comprehensive evaluation metrics. However, the reliance on synthetic data makes it potentially less suitable for our setting, because the text encoder used for the reference embeddings was not fine-tuned on the synthetic data. The Real World Knowledge Unlearning (RWKU) benchmark \cite{jin2024rwku} used in this paper evaluates real-world celebrity data and includes evaluations for robustness to adversarial attacks. Recent work \cite{scholten2025probabilistic} on evaluating unlearning has also highlighted that sampling answers probabilistically instead of using greedy decoding breaks state-of-the-art unlearning methods. 

\subsection{Adversarial Attacks}
A number of adversarial attacks have been developed that demonstrate the limited robustness of current unlearning methods. These range from black-box prompting (e.g., paraphrasing, multi-hop questions) \cite{patil2023can, zhong2023mquake} to attacks targeting the model's internal representations \cite{patil2023can, seyitouglu2024extracting}. Worryingly, even simple post-unlearning actions like quantization \cite{zhang2024catastrophic} or retraining (even on unrelated data) \cite{hu2025unlearning} can restore the supposedly unlearned knowledge. This further underscores the limitations of state-of-the-art unlearning methods. Since our method can be applied at the outputs of intermediate layers, not only the final tokens, we hypothesize that it may be more effective at counteracting some adversarial attacks that target hidden states.

\end{document}